\documentclass[letterpaper, 10pt, conference]{ieeeconf}      

\IEEEoverridecommandlockouts                              

\let\labelindent\relax

\usepackage{graphics} 
\usepackage{amsmath} 
\usepackage{amssymb}
\usepackage{subcaption}
\usepackage{graphicx}  
\usepackage{enumitem}
\usepackage{tikz}
\usepackage{amsfonts}
\usepackage{multirow}
\usepackage{dsfont}
\usepackage{float}
\usetikzlibrary{arrows,automata}
\usepackage{pgfplots}
\usepackage{capt-of,etoolbox}
\usetikzlibrary{intersections}

\title{\LARGE \bf
Multi-Head Attention for Multi-Modal Joint Vehicle Motion Forecasting
}

\author{Jean Mercat$^{1, 2}$, Thomas Gilles$^{1, 3}$, Nicole El Zoghby$^{1}$, \\%
Guillaume Sandou$^{2}$, Dominique Beauvois$^{2}$, and Guillermo Pita Gil$^{1}$
\thanks{*This work was made in collaboration with L2S and Renault SAS}
\thanks{$^{1}$ Department of data fusion,
        Technocentre Renault, 78280 Guyancourt, France
        {\tt\small \{Jean.Mercat, Thomas.Gilles, Nicole.El-Zoghby, Guillermo.Pita-Gil\}@renault.com}}%
\thanks{$^{2}$ Laboratoire des signaux et des systèmes,
        Centrale-Supélec, 91192 Gif sur Yvette, France
        {\tt\small \{Jean.Mercat, Guillaume.Sandou, Dominique.Beauvois\}@centralesupelec.fr}}%
\thanks{$^{3}$ \'Ecole polytechnique,
        Route de Saclay, 91128 Palaiseau
        {\tt\small Thomas.Gilles@polytechnique.edu}}%
}

\begin{document}

\tikzset{
    state/.style={
           rectangle,
           draw=black, very thick,
           minimum height=2em,
           inner sep=2pt,
           text centered,
           },
    name plot/.style={every path/.style={name path global=#1}}
}

\pgfmathdeclarefunction{dnorm}{2}{%
  \pgfmathparse{1/(#2*sqrt(2*pi))*exp(-((x-#1)^2)/(2*#2^2))}%
}

\maketitle
\thispagestyle{empty}
\pagestyle{empty}


\begin{abstract}

This paper presents a novel vehicle motion forecasting method based on multi-head attention.
It produces joint forecasts for all vehicles on a road scene as sequences
of multi-modal probability density functions of their positions.
Its architecture uses multi-head attention to account for complete interactions between all vehicles,
and long short-term memory layers for encoding and forecasting.
It relies solely on vehicle position tracks, does not need maneuver definitions, and does
not represent the scene with a spatial grid.
This allows it to be more versatile than similar model while combining many forecasting capabilities,
namely joint forecast with interactions, uncertainty estimation, and multi-modality.
The resulting prediction likelihood outperforms state-of-the-art models on the same dataset.

\end{abstract}

\section{INTRODUCTION}
Automation of driving tasks aims for safety and comfort improvements.
For that purpose, Autonomous Driving (AD) systems rely on the anticipation of the
traffic scene movements.
Consequently, motion forecasting is used in AD algorithms such as path planning and target selection.
The main obstacle in this task is the human driver behavior that can neither be modeled nor predicted perfectly.
It is especially challenging in negotiating situations with many participants
where drivers' interactivity plays a determinant role.
A technical challenge is to find a representation of the road scene that allows
forecasting algorithms to account for interactions within a variable number of observed
vehicles.
It should do so with an unevenly distributed observation accuracy on a wide partially occluded surrounding area.
Occlusions and uneven accuracy do not affect the closely related topic of pedestrian motion forecasting
where, in most applications, the observations are not embedded in the scene.
Two other aspects are specific to vehicles and should be considered:
the importance of the road network structure and of the reaction time due to inertia.
This requires the understanding of the road network structure and longer time and distance anticipation.

The unknown driver's decisions and the perception inaccuracies make forecasting uncertainties unavoidable.
In that context, another objective is to control the uncertainties of the motion forecasts.
The scene uncertainties are characterized with a probability density function that presents modes with dispersion.
Modes are local maxima of the Probability Density Function (PDF) of future positions.
They stand for occurrences of choices, for example, a driver chooses a lane,
or the perception system chooses a classification.
Dispersions around each mode represent the continuous uncertainties.
They are small errors made at each step of the process such as perception,
estimation, and some model approximations.
When considering multiple modes, there is a challenging trade-off to find between anticipating a
wide diversity of modes and focusing on realistic ones.
To meet the need for anticipation in AD systems, forecasting algorithms must control the
two kinds of uncertainties within an interaction aware framework.
%

\section{Related work}


\textbf{Learned models} for trajectory and maneuver forecast are compared in the survey~\cite{Lefevre2014}.
Since then, recurrent neural networks mostly using the Long Short-Term Memory~\cite{Hochreiter1997} (LSTM)
architecture have become the standard technology for statistical trajectory forecasting.
It has been used with the same kind of hand-crafted interaction features than traditional models
such as social force~\cite{Helbing1998} but it failed to generalize to complex situations.
This was overcome with the social pooling mechanism in~\cite{Alahi2016} by using a spatial grid representation.
It places features computed with LSTMs on a coarse \emph{spatial grid} to allow spatially related sequences to share features.
The subsequent work~\cite{Deo2018}, used as a baseline in our application, uses convolutional social pooling
on a coarse spatial grid.
Spatial grids are representation spaces that are able to account for a variable number of input vehicles without
ordering.
An extension of the convolutional social pooling made in~\cite{Messaoud2019} uses non-local multi-head attention
over spatial grids to account for long distance interdependencies.
Spatial grid representations limit the zone of interest to a predefined fixed size
and the spatial relation precision to the grid cell size.
To allow direct interactions without spatial grids, in~\cite{Gupta2018}, a feature-wise
max-pooling of relative positions encoding similar to the PointNet~\cite{Qi2017} architecture is used.
However, the social context treats all other agent uniformly whereas
agents should interact with a selection of relevant other agents.

\textbf{Social attention} is a mechanism that allows selective interactions within relevant agents.
It is used to make the social context more specific.
In~\cite{Sadeghian2018a} a different context vector is built for each agent.
Other agents features are sorted according to their relative distance to the target agent in a list.
This list has a fixed size of $N_{\text{max}}$ agents and is sensitive to small variations of other agents positions.
A soft attention mechanism is used on this list to produce a context feature vector.
To avoid the list ordering sensitivity,~\cite{Amirian2019} uses hand crafted relative geometric features to build
a set of normalized weights.
The context vector is a convex sum of other agent's feature vectors that is invariant to the ordering.
These last three solutions are used within a Generative Aderserial Network (GAN) architecture.
Generative models such as GANs and Variational Auto-Encoders (VAE) are able to describe complex distributions.
However, they are only able to generate an output distribution with sampling and do not express a PDF.
More complex interactions than simple distances and angles should be produced in the context of vehicle forecasting
to account for specific behaviors such as following or yielding.
This is made using a graph representation of vehicles neighbors in~\cite{Li2019}.
However it only account for local interactions among vehicles within a hand-defined distance threshold.
A dot product attention mechanism is produced in~\cite{Vemula2018}.
It is inspired from the attention mechanism first developed in~\cite{Vaswani2017} for sentence translation.
This mechanism allows joint forecast of every vehicle in the observed scenes without spatial limitations.
It accounts for long range interactions within a varying number of vehicles and does not require the ordering of the
vehicles tracks it takes as input.
In~\cite{Vemula2018}, this dot product attention is used within a spatio-temporal graph representation
of the scene developed in~\cite{Jain2016}.
This representation combines spatial and temporal dependencies that rely on
positions, pedestrian relative positions, and time step movements as features.
Each are embedded with LSTMs before using the dot product attention for social interactions.
This identifies important relations between neighbors to be considered for interactions and combines the pedestrian
feature representation with the feature representation of the relations.
In this work we use the multi-head extension of this attention mechanism, also from~\cite{Vaswani2017},
with a different road scene representation.
We do not rely on spatio-temporal graphs but on a simple temporal embedding followed by social interactions that
allows interactions between more complex feature representations than only temporal and relative dynamics.
We show in the application (section~\ref{sec_application}) that different heads specialize
to different and interpretable interaction patterns.
Our network outputs a mixture of bivariate Gaussian laws that is more adequate to describe the expected distribution than the
simple bivariate Gaussian law from~\cite{Vemula2018} and we show that it produces diversified multi-modal forecasts.

\textbf{Multi-modal} forecasts are expressed as predictions with local probability maxima.
Mixture Density Networks defined in~\cite{Bishop1994} are used in~\cite{Deo2018}
that predefines driving maneuvers as prediction modes.
Each maneuver mode is matched with one Gaussian component of the mixture and a conditional predictor is trained along
with a maneuver classifier.
As shown in~\cite{Shouno2018}, the various modes in the trajectory data are very complex and numerous.
Thus, capturing them with a few predefined maneuvers is not enough.
Using Gaussian mixture does not necessarily produce diversified modes.
A solution to obtain distinct predicted modes without pre-defining them is proposed in~\cite{Yuan2019}.
However, it changes the optimized
objective that no longer maximize the forecast likelihood.
In~\cite{Bhattacharyya2018} another solution that preserves the forecast statistics while
producing diversified predictions is proposed.
However, both methods rely on VAEs that generate prediction samples but not the PDF.
In~\cite{Cui2019}, a Multiple-Trajectory Prediction (MTP) loss is used to produce multimodal trajectory predictions
without the need for sampling.
However, as in~\cite{Yuan2019}, this modifies the objective function and alters the forecast likelihood.

\noindent
To the best of our knowledge, our added contributions are:
\begin{itemize}
    \item The use of multi-head attention for motion forecasting leading to specialized interactions.
    \item The combination of long range attention with joint and multi-modal forecasts.
    \item The unsupervised obtention of diversified multi-modal predictions by directly maximizing the forecast
    likelihood leading to improved likelihood of the results.
\end{itemize}

%
%
%
%
%
%
%

\section{Inputs and outputs}
\label{sec_obj}


\label{sec_inout}

\textbf{The inputs} are sequences of all vehicle $(x, y)$ positions in a road scene.
At each time $t_0$, we consider an observation history with a fixed observation frequency and a
fixed number of observations $n_{\text{hist}}$.
The past trajectory is written $\{(x, y)_{k}\}_{k=-n_{\text{hist}}+1, 0}$.
The coordinate system is centered on the ego vehicle position at $t_0$.

\textbf{The outputs} are $n_{\text{pred}}$ sequences of Gaussians mixtures for each vehicle.
It is expressed with a sextuplet $(\hat{x}, \hat{y}, \sigma_x, \sigma_y, \rho, p)$
for each vehicle, at each forecast step and for each mixture component.
It defines a Gaussian component $(\mathcal{N}((\hat{x}, \hat{y}), \Sigma), p)$ with
    \[\Sigma =
    \left( \begin{matrix}
        \sigma_x^2 & \rho \sigma_x \sigma_y \\
        \rho \sigma_x \sigma_y & \sigma_y^2
    \end{matrix}\right)
    \]
the covariance matrix, and $p$ the mixture weight such that for $n_{mix}$ components, $\sum_{m=1}^{n_{mix}}{p_m} = 1$.

\textbf{The forecasting model} is a set of functions $pred_\theta : inputs \rightarrow outputs$.
The inputs and outputs sets are defined with the cartesian products:
\begin{equation*}
    \begin{split}
    \text{inputs} &\in \left( \mathbb{R}^{2} \right)^{n_{\text{hist}} \times n_{\text{veh}}}\\
    \text{outputs} &\in \bigg(\Big( {
    \underbrace{\vphantom{\mathbb{R}_+^{2}}
        \mathbb{R}^{2}}_{\hat{x}, \hat{y}}} \times
    {\underbrace{
        \mathbb{R}_+^{2}}_{\sigma_x, \sigma_y}} \times
    {\underbrace{\vphantom{\mathbb{R}_+^{2}}
        [-1, 1]}_{\rho}}\Big)^{n_{\text{mix}}} \times
    {\underbrace{\vphantom{\mathbb{R}_+^{2}}
    \Delta^{n_{\text{mix}}}}_{p}} \bigg)^{n_{\text{pred}} \times n_{\text{veh}}}
    \end{split}
\end{equation*}

$\Delta^{n_{\text{mix}}}$ is the $n_{\text{mix}}$ elements simplex:
\[\Delta^{n_{\text{mix}}} = \Big\{(p_1,\dots,p_{n_{\text{mix}}})\in\mathbb{R}^{n_{\text{mix}}}~\Big|~\sum_{m = 1}^{n_{\text{mix}}} p_m = 1,~ p_m \ge 0~ \forall m\Big\}\]

The $pred_\theta$ function set is defined as a neural network with weights $\theta$.
$pred_\theta$ is equi-variant with permutations along the vehicle axis and it
is defined for all numbers of vehicles $n_{\text{veh}}$ and forecast steps $n_{\text{pred}}$.

\section{Model architecture}

This model uses an encoder-decoder structure.
It is based on LSTM networks for encoding and forecasting.
We propose to add two multi-head self-attention layers to this architecture to account for interactions.
The first attention layer is added after encoding to incorporate current time interactions.
The second attention layer is added after forecasting time unrolling.
This allows the forecast position sequences to remain coherent with each other.


\subsection{Global architecture}
\vspace{-10pt}
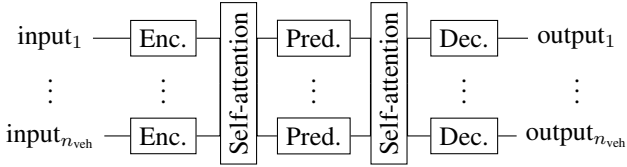
\begin{figure}[ht]
   \centering
   \vspace{7pt}
    \begin{tikzpicture}
        \node(X1){input$_1$};
        \node[below of=X1, node distance=0.6cm](X2){$\vdots$};
        \node[below of=X2, node distance=0.7cm](X3){input$_{n_{\text{veh}}}$};

        \node[draw, right of=X1, node distance=1.5cm, rectangle](ENC1){Enc.};
        \node[below of=ENC1, node distance=0.6cm](ENC2){$\vdots$};
        \node[draw, right of=X3, node distance=1.5cm, rectangle](ENC3){Enc.};

        \path (X1) edge (ENC1);
        \path (X3) edge (ENC3);

        \node[draw, rectangle, right of=ENC2, node distance=1cm](TRANS1){\rotatebox{90}{Self-attention}};

        \draw (ENC1.east) -| (TRANS1.west);
        \draw (ENC3.east) -| (TRANS1.west);

        \node[draw, right of=ENC1, node distance=2cm, rectangle](PRED1){Pred.};
        \node[below of=PRED1, node distance=0.6cm](PRED2){$\vdots$};
        \node[draw, below of=PRED2, node distance=0.7cm, rectangle](PRED3){Pred.};

        \draw (TRANS1.east) |- (PRED1.west);
        \draw (TRANS1.east) |- (PRED3.west);

        \node[draw, rectangle, right of=PRED2, node distance=1cm](TRANS2){\rotatebox{90}{Self-attention}};

        \draw (PRED1.east) -| (TRANS2.west);
        \draw (PRED3.east) -| (TRANS2.west);

        \node[draw, right of=PRED1, node distance=2cm, rectangle](DEC1){Dec.};
        \node[below of=DEC1, node distance=0.6cm](DEC2){$\vdots$};
        \node[draw, below of=DEC2, node distance=0.7cm, rectangle](DEC3){Dec.};

        \draw (TRANS2.east) |- (DEC1.west);
        \draw (TRANS2.east) |- (DEC3.west);

        \node[right of=DEC1, node distance=1.5cm](Y1){output$_1$};
        \node[right of=DEC2, node distance=1.5cm](Y2){$\vdots$};
        \node[right of=DEC3, node distance=1.5cm](Y3){output$_{n_{\text{veh}}}$};

        \draw (DEC1.east) -- (Y1.west);
        \draw (DEC3.east) -- (Y3.west);

    \end{tikzpicture}
    \caption{Block representation of our forecasting model.
    Inputs are the sequence of past observations of each vehicle.
    Outputs are the Gaussian mixture forecasts.}
    \label{sch_whole_model}
\end{figure}
\vspace{-10pt}

The figure~\ref{sch_whole_model} breaks the model in four parts: Encoder, Self-Attention, Predictor, and Decoder.
The two self-attention layers have similar architectures with different weights whereas the
encoder, predictor, and decoder use shared weights.

\subsection{Encoder}

The encoder acts as a current state estimation for each vehicle using the past observation sequences.
This state is an intermediary vector of the neural network and is difficult to interpret.
However, since it should encode the current state with at least the information of position, kinematic state,
and interaction features, it should have a sufficient dimension, we chose 128.
The input $(x, y)$ position sequences are fed to a one dimensional convolutional layer with a kernel
of size 3 sliding over the time dimension that creates sequences of 128 features for each vehicle.
This first layer increases the number of features in the vector used for the following computations.
A convolution allows this first layer to compute derivatives, smoothed values and other features extracted
from successive positions.
Then each feature sequence is encoded with a Long Short-Term Memory (LSTM)~\cite{Hochreiter1997}
into a vector of 128 features for each vehicle.

\subsection{Self-attention}
\label{sec_self_attention}

\begin{figure}[ht]
    \centering
    \vspace{5pt}
    \begin{tikzpicture}[scale=0.8, every node/.style={scale=0.8}]
        \node(X1){vehicle$_1$};
        \node[below of=X1, node distance=2cm](X2){$\vdots$};
        \node[below of=X2, node distance=2cm](X3){vehicle$_{n_{\text{veh}}}$};

        \coordinate[right of= X1, node distance=1cm](X1b){};

        \draw (X1) -- (X1b);

        \node[draw, right of=X1b, node distance=1cm](LQ1){$L_{q}$};
        \node[draw, above of=LQ1, node distance=1cm](LV1){$L_{v}$};
        \node[draw, below of=LQ1, node distance=1cm](LK1){$L_{k}$};

        \draw (X1b) -- (LV1);
        \draw (X1b) -- (LQ1);
        \draw (X1b) -- (LK1);

        \node[right of=LV1, node distance=1cm](V1){$\mathbf{v}_1$};
        \node[right of=LQ1, node distance=1cm](Q1){$\mathbf{q}_1$};
        \node[right of=LK1, node distance=1cm](K1){$\mathbf{k}_1$};

        \draw (LV1) -- (V1);
        \draw (LQ1) -- (Q1);
        \draw (LK1) -- (K1);

        \coordinate[right of= X3, node distance=1cm](X3b){};

        \draw (X3) -- (X3b);

        \node[draw, right of=X3b, node distance=1cm](LQ3){$L_{q}$};
        \node[draw, above of=LQ3, node distance=1cm](LV3){$L_{v}$};
        \node[draw, below of=LQ3, node distance=1cm](LK3){$L_{ k}$};

        \draw (X3b) -- (LV3);
        \draw (X3b) -- (LQ3);
        \draw (X3b) -- (LK3);

        \node[right of=LV3, node distance=1cm](V3){$\mathbf{v}_{n_{\text{veh}}}$};
        \node[right of=LQ3, node distance=1cm](Q3){$\mathbf{q}_{n_{\text{veh}}}$};
        \node[right of=LK3, node distance=1cm](K3){$\mathbf{k}_{n_{\text{veh}}}$};

        \draw (LV3) -- (V3);
        \draw (LQ3) -- (Q3);
        \draw (LK3) -- (K3);

        \coordinate[right of=V1, node distance=0.3cm](TOP){};
        \coordinate[right of=K3, node distance=0.3cm](BOT){};
        \draw[decorate,decoration={brace}] (TOP) -- node[left=5pt]{} (BOT);

        \node[right of=X2, text width=3cm, node distance=4.5cm](EQ){\footnotesize \[ V = \left( \begin{matrix}
                                                            \mathbf{v}_1 \\
                                                            \vdots \\
                                                            \mathbf{v}_{n_{\text{veh}}}
                                                          \end{matrix}\right) \]
                                                    \\
                                                   \footnotesize \[ Q = \left( \begin{matrix}
                                                            \mathbf{q}_1 \\
                                                            \vdots \\
                                                            \mathbf{q}_{n_{\text{veh}}}
                                                           \end{matrix} \right)\]
                                                    \\
                                                    \footnotesize \[K = \left( \begin{matrix}
                                                            \mathbf{k}_1 \\
                                                            \vdots \\
                                                            \mathbf{k}_{n_{\text{veh}}}
                                                          \end{matrix} \right)\]
                                                  };

        \node[right of=EQ, node distance=1.5cm, rotate=90](EQ2){
        \footnotesize $\text{output}=\underset{\text{dim=last}}{\operatorname{Softmax}}\left(\frac{QK^T}{\sqrt{d_k}}\right)V$};

        \coordinate[right of =EQ2, node distance=1cm](CO2){};
        \coordinate[above of=CO2, node distance=2cm](CO1){};
        \coordinate[below of=CO2, node distance=2cm](CO3){};

        \node[right of=CO2, node distance=1cm](O2){$\vdots$};
        \node[right of=CO1, node distance=1cm](O1){output$_1$};
        \node[right of=CO3, node distance=1cm](O3){output$_{n_{\text{veh}}}$};

        \draw (CO1) -- (O1);
        \draw (CO3) -- (O3);

        \draw[decorate,decoration={brace, mirror}, xshift=-2cm] (CO1) -- node[left=15pt]{} (CO3);

    \end{tikzpicture}
    \caption{Schematic representation of one attention head computations.
    Blocks $L_{q}$, $L_{v}$, $L_{k}$ are matrix multiplications of the input vectors.}
    \label{sch_selfattention}
    \vspace{-10pt}
\end{figure}
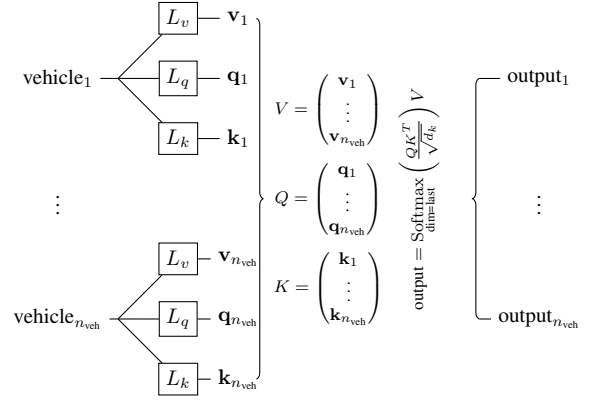

The multi-head self-attention layers allow vehicle interactions while keeping independence from their number
and ordering.
This mechanism is described in~\cite{Vaswani2017} where it is applied on sentence translation.
In this section we explain its use for vehicle interactions.
The computations made by each attention head is represented on figure~\ref{sch_selfattention}.

Each vehicle should pay attention to specific features from a selection of the other vehicles.
This is made with four steps: pulling together specific features, identifying these feature collections,
enquiring among identifiers, and gathering the results.
Each head produces a different selection of features using a linear projection of
the input tensor resulting in the value tensor $V$.
To identify these features, a key tensor $K$ is associated to each value.
Then, each vehicle must select which other vehicle to pay attention to.
For that purpose, a query $Q$ is produced to find a selection of keys.
The match score between a key and a query is their dot product, it is scaled with the square root of the key
dimension $\sqrt{d_k}$ and normalized with a softmax.
This produces an attention matrix that contains coefficients close to $1$ for matching queries and keys
and close to $0$ otherwise.
The attention matrix is square of size $n_{veh}$, each coefficient $(i, j)$ is the attention
coefficient of vehicle $i$ on vehicle $j$.
Finally, this matrix is used to gather the values from $V$.
Thus, the self-attention computation for each head is written:
\begin{equation}
    \text{output}=\underbrace{\underset{\text{dim}=\text{last}}{\operatorname{Softmax}}\left(\frac{QK^T}{\sqrt{d_k}}\right)}_{\text{attention matrix}}V
    \label{eq_selfattention}
\end{equation}
The outputs from all heads are concatenated and combined with a linear layer.
The resulting tensor is then added to the input as in residual networks.
\subsection{Predictor and decoder}
Tensors produced with the self-attention layer are repeated $n_{\text{pred}}$ times to be fed as time sequences
into a second LSTM layer called predictor.
This produces intermediary time sequences with some interaction awareness.
Within the feature sequences, vehicle interactions may depend on time.
Thus, we placed a second multi-head self-attention layer before feeding the output to the decoder.


Feature sequences are decoded with two linear layers shared for each time step and ReLU activations.
Finally, a last linear layer produces the mixture of Gaussian coefficients.
The output is described in section~\ref{sec_inout}.
Let $o_i$ be the $i^{\text{th}}$ coordinate of the output tensor before the activation function.
To constraint it, the following activation function is applied on each coordinate at every time steps:
\begin{equation*}
    \begin{split}
        &\{(\hat{x}, \hat{y}, \sigma_x, \sigma_y, \rho, p)\}_{m=1,n_{\text{mix}}}\\
        &~=\text{activation}(\{o_1, o_2, o_3, o_4, o_5, o_6\}_{m=1,n_{\text{mix}}})\\
        &~=\Big\{(o_1, o_2, e^{\frac{o_3}{2}}, e^{\frac{o_4}{2}}, \operatorname{tanh}(o_5), \underset{m \in \text{mix}}{\operatorname{Softmax}}(o_6))\Big\}_{m=1,n_{\text{mix}}}
    \end{split}
\end{equation*}

\section{Architecture discussion}

\subsection{Multi-head self-attention}

The general idea of this architecture is to use the good properties of the key-query self-attention layer to account for
interactions.
This offers flexibility to the model allowing powerful LSTM models to compute the features and predictions with
the fixed size inputs it demands while accepting a varying number of interacting vehicles without ordering.
This allows the simultaneous forecast of each vehicle in the scene with vehicle to vehicle interactions.

This method could be stated as computationally expensive because attention over $n_{veh}$ objects relies on computations
involving an $n_{veh} \times n_{veh}$ attention matrix.
However, in our use cases, the number of objects is lower than $n_{veh}=30$ and most of the time it remains around $10$.
This small number makes global attention affordable in all of our test cases.

We show in the application section~\ref{sec_application_attention} that multi-head
attention produces interpretable interactions with heads specializing on different interaction patterns.

\subsection{Maneuver free multimodal forecast}

With the present model, after defining a constant number of mixture components, they are diversified solely with
the loss minimization.
The loss is only the NLL value averaged over time.
Thus, minimizing it pushes the predicted modes toward those of the data distribution.
What is forecast is not a mixture of trajectory density functions but a sequence of position mixture
density functions.
There is a dependency between forecasts at time $t_k$ and at time $t_{k+1}$ but no explicit link between the modes
at those times.
To simplify, we assume that mixture components centers define local maxima of the probability distributions
and can be tracked in time by matching similar mixture coefficient values.
They are used as forecast trajectories.
Even if the human reasoning and some performance indicators use trajectories, only position PDFs
at each time steps are needed for the applications such as path planning and safety assessment.

\subsection{Hyperparameters}

This neural network is defined with a few specific hyperparameters that should be tuned: number of encoded features, number of
embedding and decoding layers and their activation functions, the number of heads in each self-attention layer,
the number of mixture component in the output distribution and the error covariance clipping value.
Other choices have been made and should be questioned such as the data normalization,
the use of shortcut connections with or without layer normalization, the use of LSTM layers, the use of two attention
layers and some implicit choices that may have been overlooked.
Optimizing the hyperparameters with a thorough process could bring some improvements and
help understand the model but is not a part of the present study.
In this work, only the general concept was prioritized and the hyperparameters were chosen from experience.

\section{Loss and performance indicators}
\label{sec_loss}
The model is trained with the Adam optimizer~\cite{Kingma2015} that minimizes the negative log-likelihood (NLL) loss.
The usual performance indicators for such forecasting models are root mean squared error (RMSE),
final displacement error (FDE), and NLL.
Only NLL accounts for the multi-modal aspect of the forecast, others are merely computed with the most probable trajectory.
None of the usual performance indicator is able to judge the trade-off between forecast accuracy
and diversity of the predicted modes.
Thus they are not entirely satisfactory and we also consider the Miss Rate (MR).

In the following equations, for the $i^{th}$ sequence at time $t_k$,
we note $(x_k^i, y_k^i)$ the observed positions,
$(\hat{x}_k^i, \hat{y}_k^i)$ the most probable forecast positions, and
$(\hat{x}^{*i}_k, \hat{y}^ {*i}_k)$ the forecast positions that produces
the minimum FDE.
$N$ is the number of sequences in the subset of the database on which the computation is made.

\vspace{2pt}
\noindent
\textbf{The RMSE} computation is made with equation~\eqref{eq_rmse} with
\begin{equation}
    \text{RMSE}(k) = \sqrt{\frac{1}{N}\sum_{i=1}^N{(x_k^i - \hat{x}_k^i)^2 + (y_k^i - \hat{y}_k^i)^2}}
    \label{eq_rmse}
\end{equation}
\noindent
\textbf{The FDE} values are less sensitive to large errors than RMSE.
Its computation is made with equations~\eqref{eq_fde}.
\begin{equation}
    \text{FDE}(k) = \frac{1}{N}\sum_{i=1}^N{\sqrt{(x_k^i - \hat{x}_k^i)^2 + (y_k^i - \hat{y}_k^i)^2}}
    \label{eq_fde}
\end{equation}
\noindent
\textbf{The Miss Rate} is the rate with which all proposed forecasts miss the final position by more than 2m.
\begin{equation}
    \text{MR}(k) = \frac{1}{N}\sum_{i=1}^N{{\mathds{1}}_{\sqrt{(x_k^i - \hat{x}^{*i}_k)^2 + (y_k^i - \hat{y}^{*i}_k)^2}>2}}
    \label{eq_mr}
\end{equation}
\noindent

The miss rate is lowered with the addition of relevant components and gives an indication about the trade-off between
accuracy and diversity.
The 2m threshold is not met if maneuvers such as lane changes are missed by all modes.

\noindent
\textbf{The NLL} computation, at each forecast time $t_k$, for each Gaussian component centered on $(\hat{x}, \hat{y})$,  with the forecast
error
$\mathbf{d} = (d_x, d_y) = (x - \hat{x}, y -\hat{y})$ and the forecast error
covariance defined with $(\sigma_x, \sigma_y, \rho)$ is written:
\begin{equation}
    \begin{split}
        \text{NLL}(d_x, d_y, \Sigma) =
        & \frac{1}{2}\underbrace{\frac{1}{(1 - \rho^2)}\biggl(\frac{d_{x}^2}{\sigma_{x}^2} + \frac{d_{y}^2}{\sigma_{y}^2}
        -2 \rho \frac{d_{x} d_{y}}{\sigma_{x} \sigma_{y}}\biggr)
        }_{\mathbf{d_k}^T\Sigma_k^{-1}\mathbf{d_k}}\\
        & + \ln\biggl(2\pi\underbrace{\sigma_x\sigma_y\sqrt{1-\rho^2}}_{\sqrt{|\Sigma_k|}}\biggr)
    \end{split}
    \label{eq_gauss_nll}
\end{equation}

\noindent
The time index $k$ is dropped to improve readability.
The computation of the overall NLL value for all mixture components is written:
\vspace{-8pt}
\begin{equation}
    \hspace{-1pt}\text{NLL}(d_x, d_y, \Sigma, p) = -\ln\left( \sum_{m=1}^{n_{\text{mix}}} p_m e^{-\text{NLL}(d_{x_m}, d_{y_m}, \Sigma_m)} \right)\hspace{-7pt}
    \label{eq_multi_nll}
\end{equation}

\noindent
The mean NLL (MNLL) is the average of the NLL from equation~\eqref{eq_multi_nll} over the test set.
Minimizing the NLL loss maximizes the likelihood of the data for the forecast.
However, it tends to overfit part of its output.
In~\cite{Lenz2017}, NLL overfitting has degraded the results, making the NLL value unreliable as a
performance indicator.
To avoid it, we clip the standard deviations with a 10cm minimum value.

\section{Application} \label{sec_application}

This model was implemented using the Pytorch library.
The NGSIM datasets US-101 and I-80 and its pre-processing were taken from
the published code accompanying the article~\cite{Deo2018}.
This also defines the dataset splitting into training, validation, and test sets.
Thus, a fair comparison with these results is made.
The dataset contains the tracks of all vehicle position on a road segment observed from a camera.
The pre-processing produces data that simulates observations from a given vehicle.
Each vehicle is alternatively chosen as the observing vehicle.
Its surroundings in adjacent lanes and within a 60m road segment are recorded
to produce a local road scene centered on the observing vehicle.
This road scene is tracked to produce 8 seconds sequences with all positions being recorded at a 5Hz frequency.
The 3 first seconds are used as past observations and the 5 next seconds are used as forecast supervision.

\begin{figure*}
        \vspace{5pt}
        \centering
        \begin{subfigure}{\textwidth}
          \centering
          \includegraphics[width=0.14\textwidth, clip, trim=0.49cm 5.06cm 43.5cm 0.01cm]{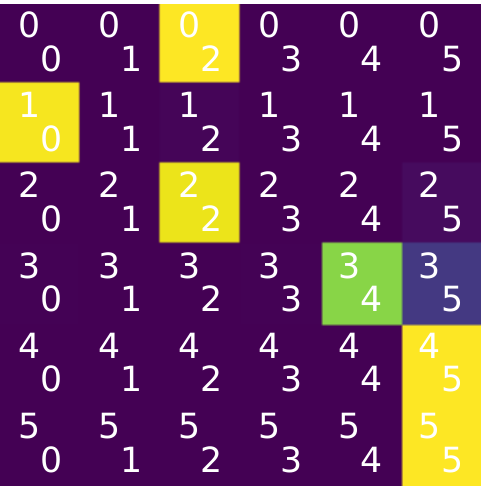}
          \includegraphics[width=0.85\textwidth, clip, trim=22cm 3.9cm 2cm 1.9cm]{attention_no_pred_131_head7.pdf}
          \caption{Head attending the front vehicle in the same lane}
          \label{fig_veh_att_a}
        \end{subfigure}\\
        \begin{subfigure}{\textwidth}
          \centering
          \includegraphics[width=0.14\textwidth, clip, trim=0.49cm 5.06cm 43.5cm 0.01cm]{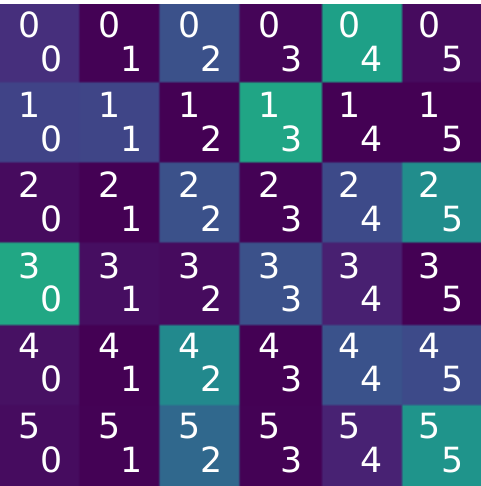}
          \includegraphics[width=0.85\textwidth, clip, trim=22cm 3.9cm 2cm 1.9cm]{attention_no_pred_131_head6.pdf}
          \caption{Head attending mainly the closest front vehicle from any lane}
          \label{fig_veh_att_b}
        \end{subfigure}\\

        \caption{
        A driving scene top view representation with all observed vehicles with their past positions in gray and the
        attention matrix for two heads of the first attention layer.
        The attention that vehicle $i$ is giving to $j$ is drawn as an arrow from $i$ to $j$, and a circle when $i=j$
        with widths proportional to the attention coefficient and a color varying with the arrow angle.
        Attention is also visible as color from purple to yellow in the $i,j$ coefficient of the matrices on the left.
        }
        \label{fig_veh_att}
        \vspace{-15pt}
\end{figure*}

\subsection{Performance indicators comparison}

\begin{table}
    \vspace{5pt}
    \centering
    \caption{Comparison of MNLL, RMSE, FDE and MR results with baselines using the same dataset.
     *CSP(M) results were recomputed with some minor modifications for a fair comparison.}
    \begin{tabular}{l|l|ccccc}
        \hline
        \multicolumn{2}{l|}{\hspace{-5pt}Time horizon}                & 1s   & 2s   & 3s   & 4s   & 5s   \\
        \hline\hspace{-3pt}
        \multirow{3}{*}{\hspace{-5pt}MNLL\hspace{-3pt}} &\hspace{-3pt}CV~\cite{Mercat2019}      & 0.82 & 2.32 & 3.23 & 3.91 & 4.46\\
                              &\hspace{-3pt}CSP(M)~\cite{Deo2018}*    &\textit{-0.41} & 1.07 & 1.93 & 2.55 & 3.08\\
                              &\hspace{-3pt}\textit{SAMMP}          & -0.36 & \textit{0.70} & \textit{1.51} & \textit{2.13} & \textit{2.64}\\
        \hline
        \multirow{3}{*}{\hspace{-5pt}RMSE\hspace{-3pt}} &\hspace{-3pt}CV~\cite{Mercat2019}      & 0.76 & 1.82 & 3.17 & 4.80 & 6.70 \\
                              &\hspace{-3pt}CSP(M)~\cite{Deo2018}*    & 0.59 & 1.27 & 2.13 & 3.22 & 4.64 \\
                              &\hspace{-3pt}GRIP~\cite{Li2019}        & \textit{0.37} & \textit{0.86} & \textit{1.45} & \textit{2.21} & \textit{3.16} \\
                              &\hspace{-3pt}\textit{SAMMP}          & 0.51 & 1.13 & 1.88 & 2.81 & 3.98\\
        \hline
        \multirow{2}{*}{\hspace{-5pt}FDE\hspace{-3pt}}&\hspace{-3pt}CV~\cite{Mercat2019}      & 0.46 & 1.24 & 2.27 & 3.53 & 4.99\\
                              &\hspace{-3pt}\textit{CSP(M)~\cite{Deo2018}*}                 & 0.39 & 0.91 & 1.55 & 2.36 & 3.39\\
                              &\hspace{-3pt}\textit{SAMMP}           & 0.31 & 0.78 & 1.35 & 2.04 & 2.90\\
        \hline
        \multirow{2}{*}{\hspace{-5pt}MR\hspace{-3pt}}&\hspace{-3pt}CV~\cite{Mercat2019}      & 0.02 & 0.20 & 0.44 & 0.61 & 0.71\\
                              &\hspace{-3pt}\textit{CSP(M)~\cite{Deo2018}*}                 & 0.004 & 0.03 & 0.12 & 0.28 & 0.44\\
                              &\hspace{-3pt}\textit{SAMMP}                                  & 0.002 & 0.02 & 0.08 & 0.15 & 0.23\\
        \hline
    \end{tabular}
    \label{tab_perf}
    \vspace{-10pt}
\end{table}


Table~\ref{tab_perf} reports results using the performance indicators defined in section~\ref{sec_loss}.
All compared models except for GRIP~\cite{Li2019} were trained and computed on the same dataset and evaluated with the same functions.
Since CSP(M) only forecasts the observing vehicle trajectory, only the errors for this vehicle are
being compared.

\textbf{Baselines}:

\emph{Constant velocity} (CV): We used a constant velocity Kalman filter with optimized parameters for forecasting
on the same data as described in~\cite{Mercat2019}.

\emph{Convolutional Social Pooling} (CSP(M)): We retrained the model from~\cite{Deo2018}.
It uses a maneuver classifier trained with preprocessed data that conditions a predictor for multimodal forecasts.
A forecast of the center vehicle trajectory is made with information from its social environment using the convolutional social
pooling mechanism.
In~\cite{Deo2018}, the model CSP with unimodal forecast gives better RMSE results
than the multi-modal forecast CSP(M).

\emph{Graph-based Interaction-aware Trajectory Prediction} (GRIP): We took the published results from~\cite{Li2019}.
It uses a spatial and temporal graph representation of the scene to make a maximum likelihood trajectory
prediction simultaneously fol all vehicles in the scene.
It produces the best results in term of RMSE but it does not account for error estimation nor multimodality.

\emph{Social Attention Multi-Modal Prediction} (SAMMP): The model described in this article.
We chose six mixture components to match the CSP(M) model for a fair comparison.


The MNLL results from table~\ref{tab_perf} show that our model improves forecast likelihood over the compared models.
The lower miss rate shows that the mixture components learned with our model are more relevant than the maneuver
definition made in the CSP(M) model and that the forecasts are well diversified.
Our RMSE results does not improve upon the results from the GRIP model when we consider the most probable trajectory.
However, the most probable trajectory is not always the closest one from the ground truth.
The RMSE of the best matching trajectory among the six output trajectories from our model (found using the ground truth)
are much lower (respectively 0.31 0.71 1.20 1.80 2.55).
Thus, the comparison with the results from GRIP are unclear because their lower RMSE could be caused by mode
averaging on their part or by an imperfect mixture probability coefficient evaluation on our part.
This could be answered by comparing the miss rates from GRIP with the miss rate of the most probable trajectory from our model.
\begin{figure}
        \centering
        \vspace{5pt}
        \includegraphics[width=\linewidth, clip, trim=5cm 5cm 0.5cm 2.5cm]{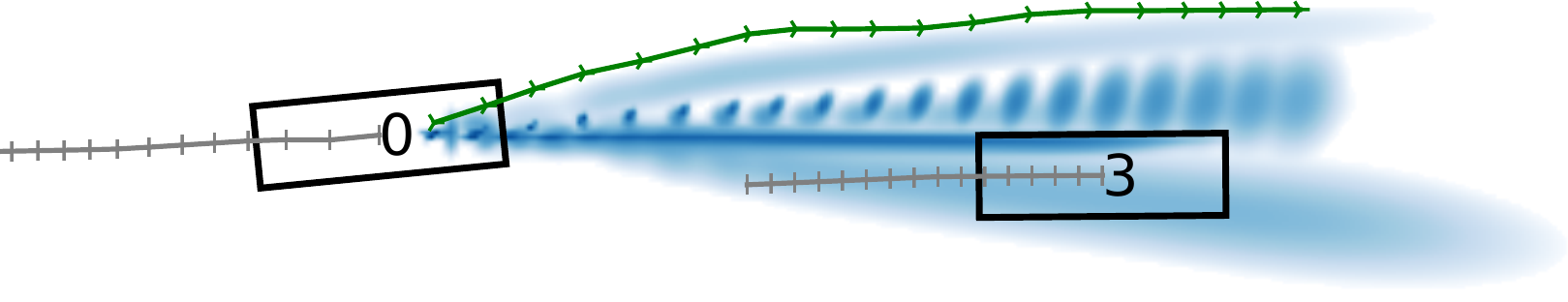}
        \caption{
        Another driving scene top view representation.
        Superposed forecast probability density functions of the ego position
        are represented in blue shades in log scale.
        Ground truth future positions are represented with a green line.
        }
        \label{fig_multimod}
        \vspace{-10pt}
\end{figure}
\subsection{Attention interpretation}
\label{sec_application_attention}
The attention matrices give insights about the importance of some interactions.
Some of the head roles can be rationalized by looking at the attention matrix it produces in different contexts.
For example, after most tested trainings of the model, one of the heads had strongly specialized for front vehicle attention
such as the one in figure~\ref{fig_veh_att_a}.
The main attention link always goes from one vehicle to the vehicle in front of it, or to itself if there is no
front vehicle.
With one training, there was not any specialized front vehicle attention head, however, one head had specialized
for rear vehicle attention.
Most of the experiments produced another head matching more or less strongly the closest front vehicle
in any lane, such as the one on figure~\ref{fig_veh_att_b}.
Other heads also specialize but the interpretation is less clear because it matches many vehicles together.
There is often a distinction between front and rear attention and a distinction of lanes.
\subsection{Multi-modal forecasting}

On the figure~\ref{fig_multimod}, the vehicle 0 aggressively overtakes the vehicle 3.
In this situation, the future holds diversified possibilities.
The overtake could be aborted or be made less aggressively, also the last observations of
acceleration and turning could be the results of perception errors.
This example has been chosen because it shows multiple lateral modes that are
easier to visualize than the more common longitudinal modes.
The NLL loss training is enough to produce a multimodal output matching those possibilities.
Using an unmodified NLL loss prevents biases in the forecast distribution that a different loss function
may cause and it indeed leads to lower NLL values.




\section{How to extend this?} \label{sec_extension}

The simplest extension is to add additional observations on each vehicle such as velocity, orientation, size or blinkers.
Another simple extension is to match various object classes such as cars, pedestrians and trucks
with specific encoders, predictors, and decoders to allow inter-class interactions.
These adaptations can easily be made because our forecasting algorithm is model-free.

Our application and the one from~\cite{Deo2018} both work with NGSIM US-101 and I-80 datasets.
They are composed of highway straight roads observed from above with a camera.
This simplistic configuration does not allow training nor testing of the road network understanding.
However, this is an important challenge that our model can be extended to consider.
In the present architecture, self-attention produces keys, queries, and values from the same input to transform the
input value.
The keys and values can be extended with additional inputs.
For example, lane center line discretized as a sequence of
position points.
A bidirectional LSTM could be a good encoder for this new kind of input.
Then, each head of the first attention layer would be extended with two additional linear layers $L_{vext}$ and $L_{kext}$,
both producing the same number of outputs and each with the same feature dimension
as the other linear layers from the attention head.
With this new input, extended keys and values would be produced :
\vspace{-10pt}
\begin{equation}
    \begin{split}
        \text{encoded}_{\text{ext}} &= \operatorname{extEncoder}(\text{input}_{\text{ext}})\\
        V_{\text{ext}} &= L_{v_{\text{ext}}}(\text{encoded}_{\text{ext}})\\
        K_{\text{ext}} &= L_{k_{\text{ext}}}(\text{encoded}_{\text{ext}})
    \end{split}
\end{equation}
\noindent
These new keys and values would simply be concatenated to the head usual keys and values:
\begin{equation}
    \text{output}=\underset{\text{dim}=\text{last}}{\operatorname{Softmax}}\left(\frac{Q[K, K_{\text{ext}}]^T}{\sqrt{d_k}}\right)[V, V_{\text{ext}}]
    \label{eq_selfattention_extended}
\end{equation}
\noindent
The output would match the vehicle features with an additional attention over the new input.

To improve the tracking of trajectories from our output, an infinite number of trajectories may be
defined as the optimal transport path between the Gaussian mixture at time $t_k$ and the one at time $t_{k+1}$.
A variable finite number of trajectories would be produced as the optimal transport paths that pass through local maxima.

\section{CONCLUSIONS}

We proposed a road scene forecasting solution that produces multimodal probability function forecast
jointly for all vehicles of the scene.
Our method generates interpretable social attention coefficients that will be extended to other road scene observations.
Results from our approach have outperformed state-of-the-art results with the NLL indicator.
This shows a good forecasting capacity as well as a good uncertainty evaluation leading to a preferred trade-off
between accuracy and prediction diversity.
Future work will include attention of vehicles to lanes and be based on the recently
published Argoverse~\cite{Chang2019} dataset in urban situations.
We expect the urban conditions to cover more complex cases and highly interactive scenes that are better suited to
show the interactive capacity of the proposed solution.

\vspace{15pt}
\subsubsection*{Acknowledgements} We are grateful to Edouard Leurent for his comments and corrections and to the
anonymous reviewers who constructively pointed out many ways to improve this paper.
\vspace{5pt}


\end{document}